\documentclass[sigconf]{acmart}
\usepackage[utf8]{inputenc}

\usepackage{lipsum}
\usepackage{amsmath}
\usepackage{fancyhdr}
\usepackage{graphicx,subcaption}
\usepackage{comment}
\usepackage{float}
\usepackage{caption}
\usepackage{tabto}
\usepackage{subcaption}
\usepackage{textgreek}
\usepackage{fixltx2e}
\usepackage{enumitem}
\usepackage{algorithm}
\usepackage{algpseudocode}

\title{Random Forest of Epidemiological Models for Influenza Forecasting}

\author{Majd Al Aawar \and Ajitesh Srivastava}

\affiliation{\institution{University of Southern California}
\city{Los Angeles}
\state{California}
\country{USA}}

\date{\today}

\begin{document}

\begin{abstract}
Forecasting the hospitalizations caused by the Influenza virus is vital for public health planning so that hospitals can be better prepared for an influx of patients. Many forecasting methods have been used in real-time during the Influenza seasons and submitted to the CDC for public communication. The forecasting models range from mechanistic models, and  auto-regression models to machine learning models. We hypothesize that we can improve forecasting by using multiple mechanistic models to produce potential trajectories and use machine learning to learn how to combine those trajectories into an improved forecast. We propose a Tree Ensemble model design that utilizes the individual predictors of our baseline model SIkJalpha to improve its performance. Each predictor is generated by changing a set of hyper-parameters. We compare our prospective forecasts deployed for the FluSight challenge (2022) to all the other submitted approaches. Our approach is fully automated and does not require any manual tuning. We demonstrate that our Random Forest-based approach is able to improve upon the forecasts of the individual predictors in terms of mean absolute error, coverage, and weighted interval score. Our method outperforms all other models in terms of the mean absolute error and the weighted interval score based on the mean across all weekly submissions in the current season (2022). Explainability of the Random Forest (through analysis of the trees) enables us to gain insights into how it improves upon the individual predictors.

\end{abstract}
\maketitle
\section{Introduction}
Contagious illnesses like the Influenza have a significant impact on global health, hospitalizing hundreds of thousands of people in the US alone on a yearly basis \cite{CDCBurden}. Accurately forecasting the hospitalizations caused by the Flu is an essential task since it enables public health officials to be better prepared and to allocate the necessary resources accordingly. It is also essential that they are provided with probabilistic forecasts which act as uncertainty measures for the decision-making process \cite{ijerph17041381, probForecasts}. 

To improve the science and usability of epidemic forecasts, the Centers for Disease Control and Prevention (CDC) facilitates open forecasting projects through the Epidemic Forecasting Initiative~\cite{EPI}. Influenza forecasting efforts began as early as 2013 and are currently ongoing with its FluSight challenge, which has numerous teams participating in it. They provide national and state level, point and probabilistic forecasts of the weekly number of influenza hospitalizations for the next 4-weeks after the forecast date \cite{CDCBurden,EPIFluSight}. The forecasting models used for the submissions included mechanistic models, auto-regressive models, machine learning models, and ensembled models \cite{FluSightData}. Examples of the mechanistic models used include both SEIRS (Susceptible - Exposed - Infectious - Recovered - Susceptible)  and SLIR (Susceptible - Latent - Infected - Removed) compartmental models, which are both commonly used in epidemiological forecasting \cite{Shaman2012,Kuddus2021}.
Such models make assumptions about the dynamics of the infectious disease and distribution of parameters that may not fully reflect the reality and thus deviate from the ground truth.
Machine learning has also been used in epidemiological forecasting \cite{Alali2022,Kamarthi2021,DeepCovid,EpiDeep} and is primarily focused on the use of deep sequential models. Machine learning models may be prone to overfitting and difficult to interpret. 

We hypothesize that we can generate multiple potential forecasts from mechanistic models and use an interpretable machine learning algorithm to combine the forecasts to yield improved results. In this way, we can leverage domain knowledge through the mechanistic models and learn how to infer forecasts from a reasonable set of potential forecasts with machine learning. 
We propose using variations of the SIkJ$\alpha$ model~\cite{Baseline:1} as mechanistic predictors of hospitalizations. Each predictor represents a different forecast generated by changing the set of hyper-parameters of the SIkJalpha model. A Random Forest is used as an ``ensembling'' technique to combine these predictors. 
We also examine Least Square Boosted Trees (LSBoost) as an ensembling method and compare our methods to all other approaches used in the current FluSight challenge (2022) and the top performing methods in the previous three seasons FluSight ILI forecasting. 
A key feature of our approach is that it is an \textit{automated approach with no human intervention}, i.e., we have an automated script that pulls the relevant data and generates the forecast. This feature makes our approach scalable as we are not relying on human expertise to tune the model for each region for which forecast is desired. 
We participated in the ongoing FluSight challenge with a similar philosophy, where all the submissions' results were generated by our scripts each week. No model tuning or state-specific tuning was performed manually.
The main contributions are the following:
\begin{itemize}
    \item We propose a Tree Ensemble (Random Forest/LSBoost) based framework which takes predictions from multiple mechanistic models and some region-specific data as input and outputs improved forecasts.
    \item We modify the SIkJalpha model for hospitalization forecasts for Influenza datasets and generate its variations to act as input predictors into the Tree Ensemble algorithm.
    \item We conduct experiments and demonstrate that our approach improves upon the input predictors.
    \item We also demonstrate that results from our approach, while being fully automated, outperforms all other submissions of the current season 
    \item We conduct ablation studies to show the impact of various modeling choices and additionally present evaluations on hospitalizations with influenza-like-illnesses for the past seasons.
    \item We deploy our approach through automated scripts (without any state-specific manual tuning or human intervention) in real-time for prospective forecasting in 2022, where our approach outperforms the ensemble of all approaches (FluSight-Ensemble).
    \item We analyze the individual trees of the Random Forest to gain insights into how the individual predictors are used to generate the final forecasts.
\end{itemize}

\section{Background}
\subsection{FluSight Challenge}

The CDC has coordinated collaborative epidemiological forecasting efforts on various diseases, including COVID-19, the West Nile Virus, Aedes~\cite{EPIFluSight}. It has also been coordinating influenza forecasting challenges, FluSight, for weekly hospitalizations since 2016. The 2016 season had teams submit national level forecasts, while later seasons had teams forecast at both the state and national level. Teams were required to submit both point and probabilistic forecasts, as well as seasonal targets based on the data collected from the  Outpatient Influenza-like Illness Surveillance Network, ILINet \cite{FluSight:2}, for the 2016 to 2019 seasons. The current FluSight2022 challenge also has teams providing point and probabilistic forecasts, at both the state and national levels, but using the data collected COVID-19 Reported Patient Impact and Hospital Capacity by State Timeseries dataset (CPIHC) \cite{FluSight:1}. These forecasts are of the weekly number of confirmed influenza hospitalizations for up to 4 weeks after the forecast due date using the specification of epidemiological weeks, which run through Sundays to Saturdays. No seasonal targets were required for this years challenge.

\subsection{Related Work}

Past and current influenza forecasting efforts have seen many methods relying on mechanistic models such as SIR (Susceptible – Infected – Recovered) compartmental models. Ben-nun et al. \cite{Ben-Nun2019} implemented a deterministic SIR modeling framework which accounts for various transmission mechanisms, such as climate conditions, coupling between different regions, and school vacations. They utilized the ILINet dataset, the NASA NLDAS-2 for specific humidity values, and approximated state school schedules from various data sources. To account for coupling, the force of infection was defined to be a function of a mobility matrix as well as the risks of infection from those within the same region j, from the infected from region i who traveled to region j, and from going from region j to region i. The specific heat values and the school schedules were used to modify the transmissibility during a time period. Parameter posteriors were determined using a Metropolis-Hastings Markov Chain Monte Carlo (MCMC) procedure. It was found that the best forecasts made use of the coupling feature but also required some level of human intervention.

Another approach taken by teams is the use of Bayesian models. Osthus et al. \cite{Dante2019} implemented Dante, a Bayesian flu forecasting model, which was used in the previous FluSight challenges and also makes use of ILINet. It consists of 2 components, the fine-scale (state) model and the aggregation (national) model. The state model assumes that the observed proportion of hospitalizations has a Beta distribution dependent on the unobserved proportion of hospitalizations. It also models the unobserved true proportion of hospitalizations as a function of four components, each one modeled as a random walk aimed to capture the common structure across the whole data -- across all state-level data, across all seasonal data, and across season/state specific data. This captures the patterns and week-to-week correlations. Inference is generated by sampling from the posterior distribution using MCMC, resulting in a sample of M draws that summarize the posterior distribution. The national-level model simply computes the forecasts as linear combinations of the state forecasts, where weights are proportional to the population estimates. This method was found to be one of the top performing methods in the 2018/2019 FluSight Challenge. Unlike existing mechanistic and Bayesian approaches, our approach relies on fast and simple mechanistic models without any assumption on the distribution of parameters and without the need for any simulations. The prediction of these models are then improved upon by a Tree Ensemble.

Deep learning approaches have also been used for forecasting but have several limitations when generating probabilistic forecasts. Kamarthi et al. \cite{Kamarthi2021} devised a neural process model, EPIFNP, which is currently being used in the FluSight2022 challenge to address these shortcomings. In addition to the ILINet dataset, it makes use of demographic, mobility, and symptomatic data. First, the input data is encoded using a deep sequential model and a latent embedding is modeled as a Gaussian random variable using a Variational auto-encoder. A stochastic data correlation graph then captures the dependencies between the reference and training sequences, thus allowing for uncertainty estimation. The final predictive distribution is parametrized with three variables, a global stochastic latent variable shared by all sequences, a local stochastic latent variable which captures the data correlation uncertainty from the correlation graph, and the stochastic sequence embeddings. These are then fed into a final single layer of a neural network designed to output the mean and variance of a Gaussian distribution. Instead of a fully data-driven model, our approach combines the advantages of mechanistic model and data-driven models. Our Tree Ensemble provides a natural way to quantify uncertainty.

\section{Methodology}\label{sec:method}
\subsection{The SIkJ$\alpha$ Model for Influenza}
\label{sec:sikja}
SIkJ$\alpha$ \cite{srivastava_fast_2020} is a heterogeneous infection rate model which can be expressed as a system of linear models for learning the parameters. 
The key idea behind the SIkJalpha model~\cite{srivastava_fast_2020} is to approximate the disease dynamics as a discrete heterogeneous rate model. For instance, suppose the new infections are created by the following true dynamics:
\begin{equation}\label{eqn:gen}
\Delta I(t) = S(t)\sum_\tau \beta'(\tau)\Delta I(t-\tau) \approx \sum_{i=1}^k \beta_i \left(I\left(t-(i-1)J\right)-I(t-iJ)\right)
\end{equation}
i.e., we approximate the dependence on the past infections by $k$ temporal bins of size $J$. Having $J>1$ has a smoothing effect. 
However, the true number of Influenza cases are not observed. Instead, we observe the hospitalizations. We extend the model to represent hospitalizations as a function of past hospitalizations. We start with the following model of hospitalizations:
\begin{equation}\label{eqn:hosp0}
\Delta H(t) = \sum_{\tau_2=0}^a \gamma(\tau_2)\Delta I(t-\tau_2)\,,
\end{equation}
i.e., infections are converted to hospitalizations for some function $\gamma(\tau)$ which when combined with Equation~\ref{eqn:gen} results in a further approximation:
\begin{align}\label{eqn:hosp_derive}
\Delta H(t) &= \sum_{\tau_2=0}^a \gamma(\tau_2)S(t-\tau_2)\sum_\tau \beta'(\tau)\Delta I(t-\tau_2 - \tau).
\end{align}
Assuming $\lvert S(t-\tau_2) - S(t) \rvert \leq \epsilon, \forall \tau_2 \in [0, a]$, we have
\begin{align}\label{eqn:hosp}
\Delta H(t) &\leq (1+\epsilon) S(t)\sum_{\tau_2=0}^a \gamma(\tau_2)\sum_\tau \beta'(\tau)\Delta I(t-\tau_2 - \tau) \nonumber \\
&= (1+\epsilon) S(t)\sum_\tau \beta'(\tau)\sum_{\tau_2=0}^a \gamma(\tau_2)\Delta I(t-\tau_2 - \tau) \nonumber \\
&= (1+\epsilon) S(t)\sum_{\tau} \beta'(\tau)\Delta H(t-\tau)
\end{align}
And, similarly,
\begin{equation}
    \Delta H(t) \geq (1-\epsilon) S(t)\sum_{\tau} \beta'(\tau)\Delta H(t-\tau)\,.
\end{equation}
Assuming a small $\epsilon$, we get the following model of hospitalizations and using the binning approach as in Equation~\ref{eqn:gen}:
\begin{equation}\label{eqn:delH}
    \Delta H(t) = S(t)\sum_{i=1}^k \beta_i \left(H\left(t-(i-1)J\right)-H(t-iJ)\right)\,.
\end{equation}
The susceptible fraction of population is approximated as
\begin{equation}\label{eqn:S}
    S(t) = 1 - \frac{I(t)}{N} = 1 - \frac{H(t)}{\mu N} \,.
\end{equation}
Here, $\mu$ is approximately the hospitalization rate which is related to $\gamma$ in Equation~\ref{eqn:hosp0} as $\mu = \sum_{\tau_2} \gamma(\tau_2)$. We treat $\mu$ as a hyperparameter within a range $(1/150, 1/50)$ that seems reasonable based on the CDC's estimate of true prevalence of Influenza in the past seasons~\cite{CDCBurden}. 

The ``alpha'' in SIkJalpha represents the learning technique. We use a weighted linear regression to emphasize recently seen data to adapt to dynamically changing behavior. Suppose we have observed the reported hospitalizations for T time steps. Then the regression is performed by minimizing the weighted least square error
\begin{equation}
LSE = \sum_{t=1}^T \alpha^{T-t} \left( \Delta H_{obs}(t) - \Delta H_{fit}(t)\right)^2\,.
\end{equation}

For generating forecasts, we recursively use Equations~\ref{eqn:delH} and~\ref{eqn:S} with the parameters identified from the regression. To capture uncertainty in the future, we assume that the future rates $\beta'$ can also be those that are seen not just this week $\lambda=0$, but those seen last week $\lambda=7$ (retraining with a week old data), which introduces another hyper-parameter along with $\mu$ that we pick from $\{1/50, 1/100, 1/200\}$. The hyperparameters $k$ and $J$ are set to 3 and 7, respectively.
Generation of quantiles for submission to the FluSight challenge using this model is described in Section~\ref{sec:deploy}

\subsection{Generating the Predictors}
\label{sec:gen_pred}

We consider three main hyperparameters which we vary to generate variations of the base model to get various trajectories. 
The weighing scheme for least squares is determined by the hyper-parameter $\alpha \le 1$, which puts more weight on the recent hospitalization trend when fitting the model. A lower value for $\alpha$ means we are putting more weight on the more recent data. The next hyper-parameter considered is the lag coefficient $\lambda$, which represents which among the recently seen rates $\beta_i$ we choose to project into the future. The final hyper-parameter considered was the reporting probability, $\mu$ which determines the magnitude of under-reporting being considered, and is multiplied by the cumulative infections when finding the number of susceptible individuals.
We end up with 30 different predictors, one for each of our selected combinations of hyper-parameters.

\subsection{Generating the Forest}
\label{sec:gen_forest}
\subsubsection{Random Forest}
\begin{figure}[!ht]
\captionsetup[figure]{justification=centering}
\centering
\includegraphics[width=0.5\textwidth]{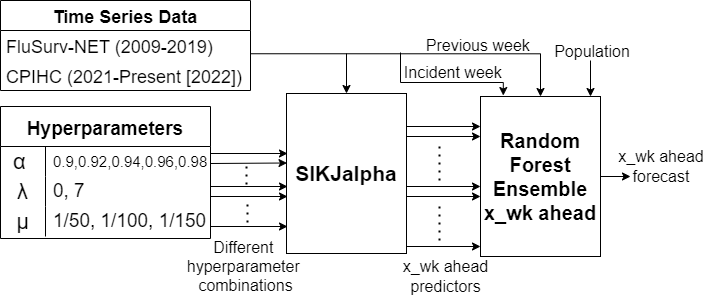}
\caption{Tree-Ensemble Design for Influenza Forecasting}
\label{fig:RFDesign}
\end{figure}
For the design of our Random Forest ensemble, we utilize the predictors generated from our baseline model. We feed it an input vector consisting of the data points corresponding to last week's hospitalizations, incident hospitalizations, the state's population size, and each individual  predictor's forecast. The time series data is normalized first with respect to the state population. The procedure is summarized in Figure \ref{fig:RFDesign}.
A total of four models are trained, each with the purpose of predicting $x$-weeks ahead, where $x = 1,2,3,$ and $4$. The number of decision trees used in the Random Forests was selected through careful tuning by first starting with 64 trees as suggested in \cite{RF:2}. The final number used was 56 since increasing the number of trees further caused a minimal change in results at the expense of larger computation costs.
The final results are the mean of each forest's individual trees' outputs for point forecasts, and the generated quantiles from the set of trees for probabilistic forecasts. 

\subsubsection{Least-Squares Boosting} We also implement LSBoosted Trees since gradient boosting can result in better performance than Random Forests given a proper choice of its parameters. The Design for our LSBoosted Trees follows the same design of the Random Forest as seen in figure \ref{fig:RFDesign} but with a change in the Tree Ensemble model. This ensembling technique uses Least-squares boosting (LSBoost) to boost 100 regression trees. The final output produced is a point forecast and can not be used to generate probabilistic forecasts. 

Additionally, we also produce a mean ensemble of all the predictors and label it ``SIkJalpha''.

\subsection{Training Procedure}\label{sec:train}
\begin{algorithm}\label{alg:train}
\caption{Training the Forest for x-weak-ahead Forecasting}\label{alg:train}
\begin{algorithmic}
\State $T \gets$ current week
\For{$t \in \{$all weeks of past seasons, $T$ weeks of this season$\}$}
\For{$s \in$ regions}
    \State $m \gets 0$
    \For{each hyperparameter $\mu, \lambda, \alpha$}
        \State $X(n, m) \gets$ x-wk-ahead prediction with SIkJalpha with $\mu,\lambda,\alpha$ trained on data up to $t-x$
        \State $m \gets m+1$
    \EndFor
    \State $X(n, :) \gets $ concatenate$(X(n, :)$, additional features of  $s)$
    \State $Y(n) \gets$ x-wk-ahead ground truth
    \State $n \gets n+1$
\EndFor
\EndFor
\State Train Tree Ensemble$(X, Y)$
\end{algorithmic}
\end{algorithm}
Algorithm~\ref{alg:train} shows the training approach.
We first train the SIkJalpha predictors for each week using only the ``observed" data seen until that week -- each season, we iterate through all that season's weekly data and train the predictors only on the weeks which we have already iterated through (hence, the observed data). We use a minimum of 4 weeks of observed data in order to have sufficient points to start the training process. Using the strategy above, we set a forecast horizon of 28 days. Therefore we get 30 predictors each of length 28 from where we can extract x\-week ahead forecast (e.g. 2 week ahead forecast is the sum of daily predicted hospitalization from day 8 to day 14 in the future). We then accumulate the predictors for each week of every season to be saved for later use in our ensemble models.
Using the generated predictors, we train our ensemble models and start the evaluation starting from the 2017 season up to the most recent 2022 season from the ILINET and CPIHC datasets, so that we generate the model predictions corresponding to each week for proper evaluation of the techniques. Note that the evaluation is performed in a rolling fashion to emulate prospective forecasting. For example, on a given season, at week $T$, $x$-week ahead task would require prediction for $T+x$. Training will be performed up to $T$ as described in Algorithm~\ref{alg:train}. This is procedure is repeated for all the weeks and seasons.

\section{Experiments}\label{sec:exp}
Our goal is to show that  (1) random Forest-based approach outperforms a naive ensemble generated by taking the mean of all the predictors (referred to as SIkJalpha in the results) retrospectively in the past seasons and in real-time in the current season, (2) our approach outperforms the individual predictors, and (3) our approach performs on par with top performing methods in the current season, despite the lack of human intervention.
\subsection{Data}\label{sec:data}
Three main data sources were used for the training and evaluation of the different models:

    \noindent \textbf{FluSurv-NET} \cite{FLUSRUV:1} is a surveillance network of hospitals across 14 states which have collected data, both demographical and clinical, on laboratory confirmed influenza hospitalizations throughout the flu seasons 2009-2022 (present day). The past hospitalization data from the 2009-2019 flu seasons were used to train our models for the FluSight 2022 challenge. The 2020-2021 season was excluded because the recorded hospitalizations were unlike any of the prior seasons and were at record lows, this was likely due to the various COVID-19 prevention measures and because no FluSight Challenge took place that year. In order to generalize this data for each state, we first generalize it across the whole US. This is done by dividing the total number of recorded hospitalizations at a given week by the ratio of the sum of populations for the 14 states to the total US population. After getting this estimation, we find the state-level data by distributing the nation-level data according to the ratio of the state-level population to the national population.
    
    \noindent \textbf{ILINET} \cite{FluSight:2} was the dataset used in the previous FluSight challenges from 2016-2020. It consists of a network of outpatient healthcare providers across all the states which report the number of patient visits for Influenza-Like-Illnesses (ILI) by age group from 2010-2022 (present day). This past data was initially used to train our models for the 2022 FluSight since it provided more data points. But it was later excluded, as mentioned in section~\ref{sec:deploy}, since Influenza hospitalizations are a subset of ILI hospitlizations and would cause our models to overestimate the forecasts. Therefore this dataset is used separately to train and evaluate our models on past FluSight challenges for ILI hospitalizations. 
    
    \noindent The \textbf{CPIHC} \cite{FluSight:1} dataset is used as the ground truth data for the forecasts in the FluSight 2022 challenge. It provides state-level data for hospitalizations collected from HHS TeleTracking, reports from health departments, and the National Healthcare Safety Network. It is also worth noting that the reporting of the flu fields became mandatory as of 02/02/2022. 
    

\subsection{Evaluation Metrics}\label{sec:metrics}
The first metric is used to evaluate our point forecasts and it is the Mean Absolute Error (MAE). The MAE is defined as:
\[MAE = \frac{1}{N}\sum_{k=1}^{N} |\hat{y_k}-y_k|\]
Where N = total number of samples being tested, $y_k$ is the target output value, and $\hat{y_k}$ is the predicted output value. Lower MAE is desirable.

The next metric used is the Weighted Interval Score (WIS)~\cite{bracher2021evaluating} to evaluate the performance of our probabilistic forecasts. The WIS is defined as:
\[WIS = \frac{1}{2K+1}\sum_{k=1}^{2K+1}2*(\{y\leq q_{\tau k}\}-\tau_k)*(q_{\tau k} - y)\,,\]
where $2K+1 =$ our total number of quantiles, $y$ is our target output value, $\tau_k$ is the kth quantile, $q_{\tau k}$ is the kth quantile prediction, and \{\} is an indicator function which returns 1 if the condition is true or 0 otherwise. Lower WIS is desirable.

The final metric used which also assess the performance of our probabilistic forecasts is the coverage~\cite{christoffersen1998evaluating}. An x\% coverage measures that the percentage of data points contained within the x\% confidence interval (CI). It is defined as:
\[Coverage = \frac{\sum_{k=1}^{N}\{lower bound\leq y_k\leq upper bound\}}{N}\]
Where N = total number of data points, $y_k$ is our target output, \{\} is an indicator function defined previously, the lower bound of our CI is the (0.5 - CI/2) quantile's prediction, and  the upper bound of our CI is the (0.5 + CI/2) quantile's prediction. Higher coverage is desirable. Ideally, $x\%$ coverage will be expected to be $x$.

The WIS and coverage evaluations are not applicable to our LSBoosted ensemble since it only generates point forecasts. 

\begin{table}[!ht]
\centering
 \includegraphics[width=0.5\textwidth]{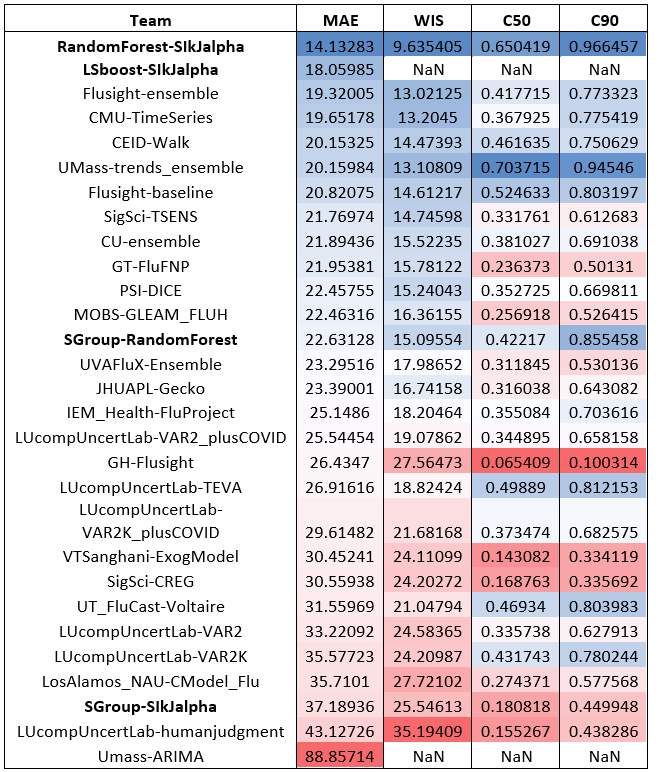}
 \caption{2022 Mean evaluations over all targets with heatmap. Blue is better, red is worse. RandomForest-SIkJalpha and LSBoost-SIkJalpha are added retrospectively. All other approaches including our two submissions SGroup-SIkJalpha and SGroup-RandomForest where submitted in real-time.}
 \label{tab:2022mean}
\end{table}

\begin{figure*}[!ht]
\centering
     \includegraphics[width=\textwidth]{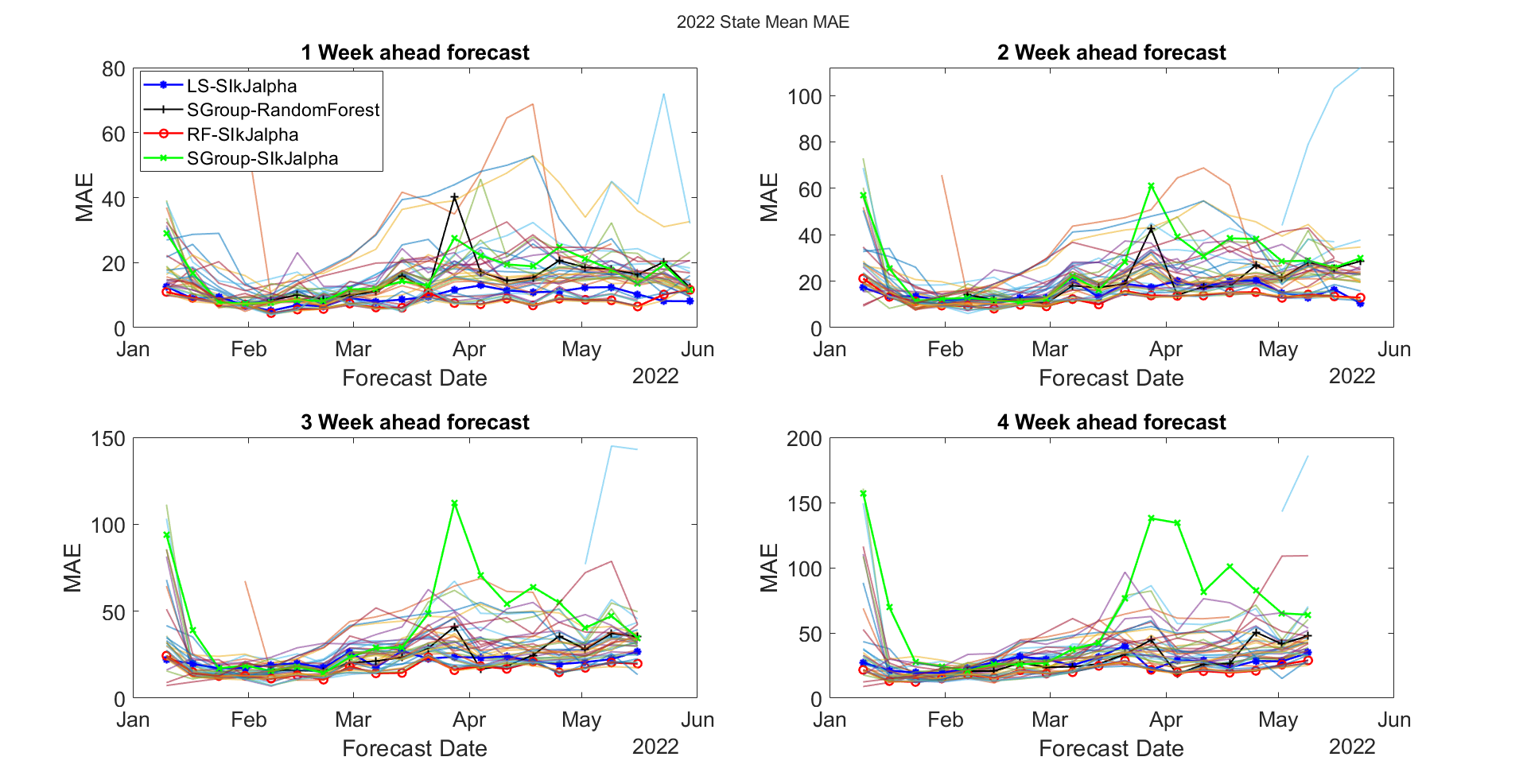}
\caption{MAE scores over time for the 2022 Flu Season showing our approach along with the our deployed versions. Blurred lines in the background represent errors from submitted forecasts from other teams.}
\label{fig:2022plot}
\end{figure*}

\begin{figure*}[!ht]
\centering
 \includegraphics[width=\textwidth]{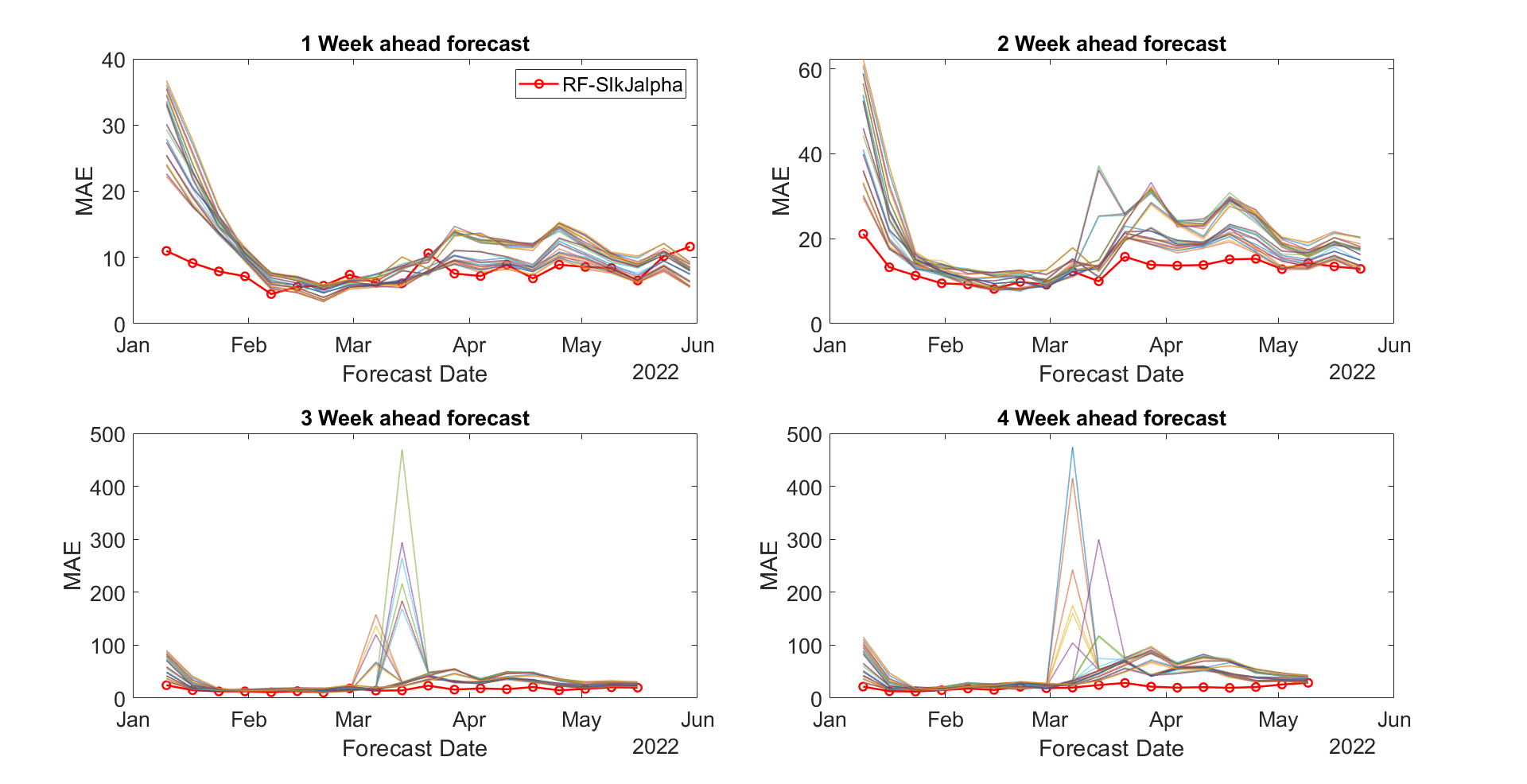}
 \caption{MAE scores over time for the 2022 Flu Season showing Individual Predictors vs RandomForest}
 \label{fig:22PvsRF}
\end{figure*}

\subsection{Results}
Each model was evaluated separately on each state and on 1, 2, 3, and 4-week ahead using MAE, WIS, and Coverage at 50\% and 90\%. We report the average performance across all these targets. Note that Coverage is bounded, while MAE and WIS are not. Typically, MAE and WIS are expected to be higher for longer horizon predictions, so the average is biased towards longer horizon performance. We believe this is reasonable as accurate longer horizon predictions provide more time for preparedness.
For the current season, we deployed two of our methods -- SGroup-SIkJalpha and SGroup-RandomForest. Details of deployment and the differences from the Random Forest-based approach discussed so far are presented in Section~\ref{sec:deploy}. Additionally, we retrospectively evaluated our RandomForest-SIkJalpha and LSBoost-SIkJalpha (described in Section~\ref{sec:method}). The results are shown in Table~\ref{tab:2022mean}.
We note that RandomForest-SIkJalpha outperformed all other submissions in terms of MAE and WIS, including both LSBoost-SIkJalpha and SGroup-SIkJalpha. The coverage at 50\% and 90\% is also high, performing 2nd best and best respectively in those fields. Figure~\ref{fig:2022plot} shows the MAE over time. We observe that RandomForest-SIkJalpha also consistently produces low errors and outperforms the individual predictors, as seen in Figure~\ref{fig:22PvsRF}.

We also observed that the RandomForest model outperformed the LSBoost model. In random forests, all the trees are generated independently through bagging. In boosting, all trees rely on the residual of the trees before which allows them to learn from the errors produced by the previous trees. Boosting is meant to reduce the bias of the final prediction, while the RandomForest is meant to reduce the variance~\cite{hastie_09_elements-of.statistical-learning}. Since we are working with daily data, which tends to be noisier than weekly data, Boosting is more likely to overfit to the noise which is likely the cause for RandomForest outperforming it.

\section{Discussion}
\subsection{In-use Deployment}\label{sec:deploy}
We participated in FluSight challenge of 2022 (Jan 10, 2022 - June 20, 2022 ) with two models. One based on the base model called SGroup-SIkJalpha without the Random Forest, and another with the Random Forest called SGroup-RandomForest. These models are described below.

\subsubsection{SGroup-SIkJalpha}
We fix the weight $\alpha$ to $0.98$ in the implementation. We vary the hyperparameter $l \in \{0, 1\}$ suggesting that the future rates may reflect the rate of either this week or the last week, and the hospitalization to case ratio hyperparameter $\mu \in \{1/50, 1/100, 1/150\}$.
For each setting of the hyper-parameter, we estimate the 95\% confidence interval of $\beta_i$. In this confidence interval, we draw 10 samples uniformly per $\beta_i$. This along with combinations of $\mu$ and $l$, result in $60$ trajectories each containing $4$ points corresponding to 1-4 weeks-ahead forecast, respectively. We generate the mean and the required 23 quantiles for our submission. All submissions are generated by an automated script run on Sunday mornings without human interventions. No manual tuning is performed based on the results. There has been only one edit to the scripts during the season. In late January, the code was adjusted to account for percentage of hospitals reporting. For instance, if on a given week for a given state, the reported hospitalization was $\Delta H(t)$, and the percentage was $P$, then the true hospitalization was assumed to be $\frac{100}{P}H(t)$. Before this adjustment, the error was high as seen in Figures~\ref{fig:2022plot} in January. Further, in April, the model projected an increase in hospitalizations. While the true hospitalizations also increased, the magnitude predicted by SGroup-SIkJalpha was much higher, resulting in high errors. These high errors towards the end have also skewed the evaluation against this model. While we noticed the discrepancy in April, we did not change or tune the approach in order to continue testing the model without human intervention. This would also provide an opportunity to clearly demonstrate the utility of the Random Forest based approach.

\subsubsection{SGroup-Random Forest}
Initially, the submission of SGroup-Random Forest only involved generating predictors from the hyper-parameters $l$ and $\mu$ (six predictors) along with the current and last week hospitalizations and population. The forecasts from this model were also generated with automated scripts with no human intervention with the following two exception. First, when the data on percentage hospitals reporting was integrated in late January to SGroup-SIkJalpha, same adjustment was made to the predictors in the Random Forest. Second, in April, we also added variations in the hyperparameter $\alpha\in \{0.90, 0.92, 0.94, 0.96, 0.98\}$. This resulted in 30 predictors generated from variations of $SIkJalpha$. Finally in late May we decided to exclude the ILINET data from our training set for the 2022 season, which resulted in the final model we used for the retrospective evaluations. It should be noted that despite poor performance of SGroup-SIkJalpha in April and May, SGroup-RandomForest was able to maintain good performance.

\subsection{Insights from the Forest}
\begin{figure}[!ht]
\centering
 \includegraphics[width=0.5\textwidth]{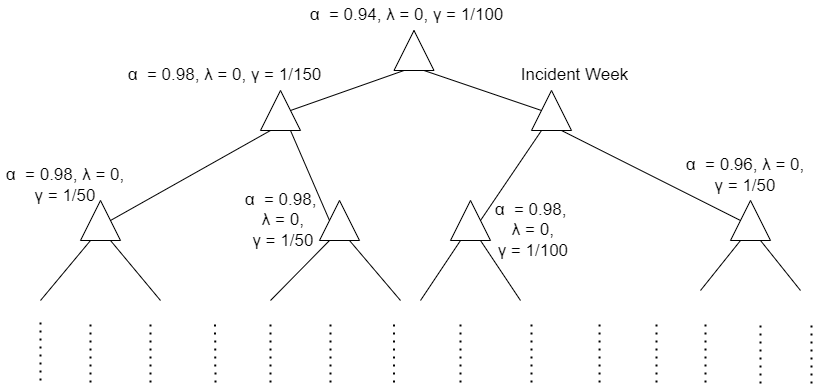}
 \caption{Top portion of sample decision tree from the 4 week ahead RandomForest model}
 \label{fig:DTSplits}
\end{figure}

\begin{figure}[!ht]
\centering
    \begin{subfigure}{\columnwidth}
     \includegraphics[width=\columnwidth]{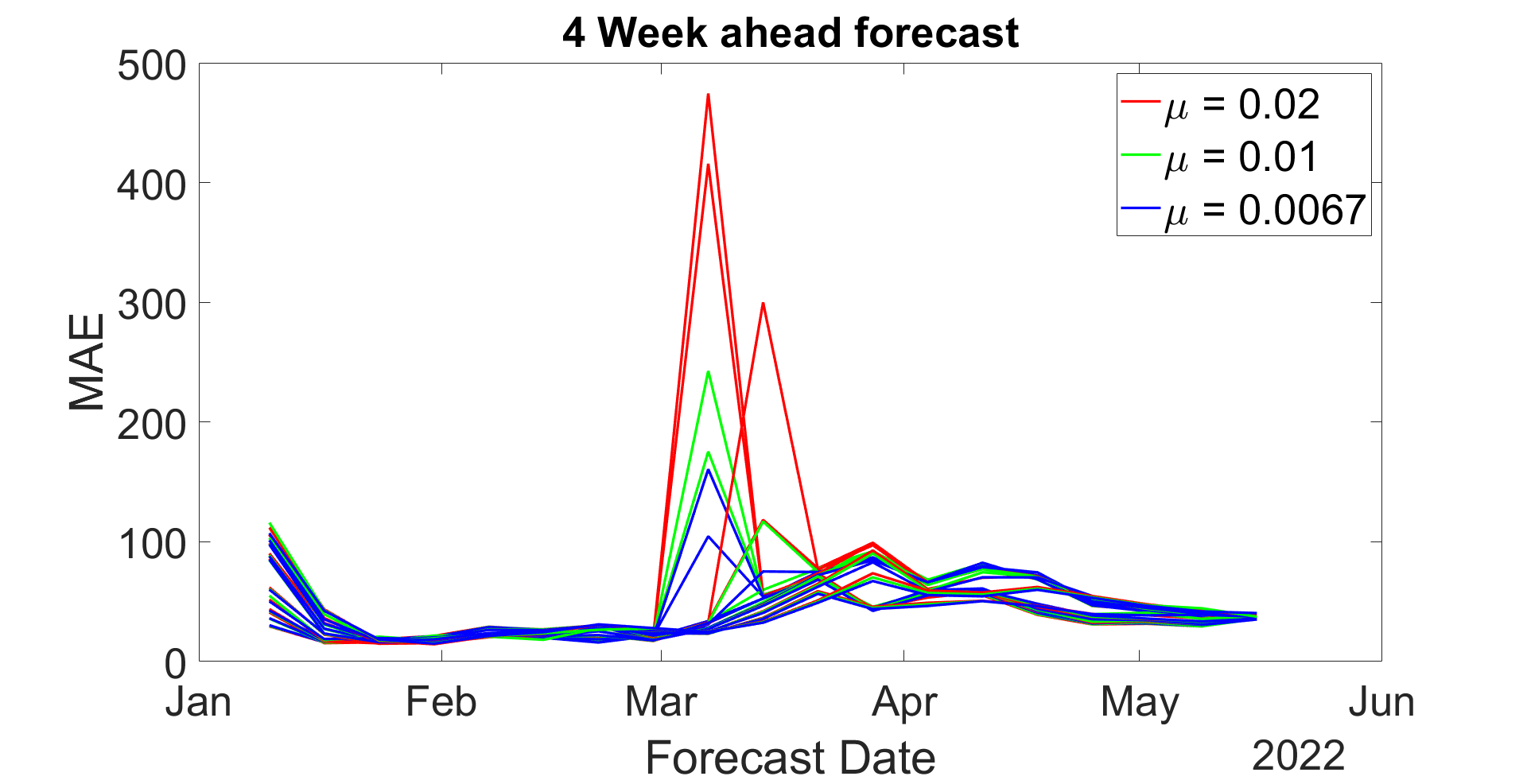}
     \caption{Sensitivity to the hospitalization to infection ratio $\mu$}
     \label{fig:uncA}
    \end{subfigure}
     \begin{subfigure}{\columnwidth}
     \includegraphics[width=\columnwidth]{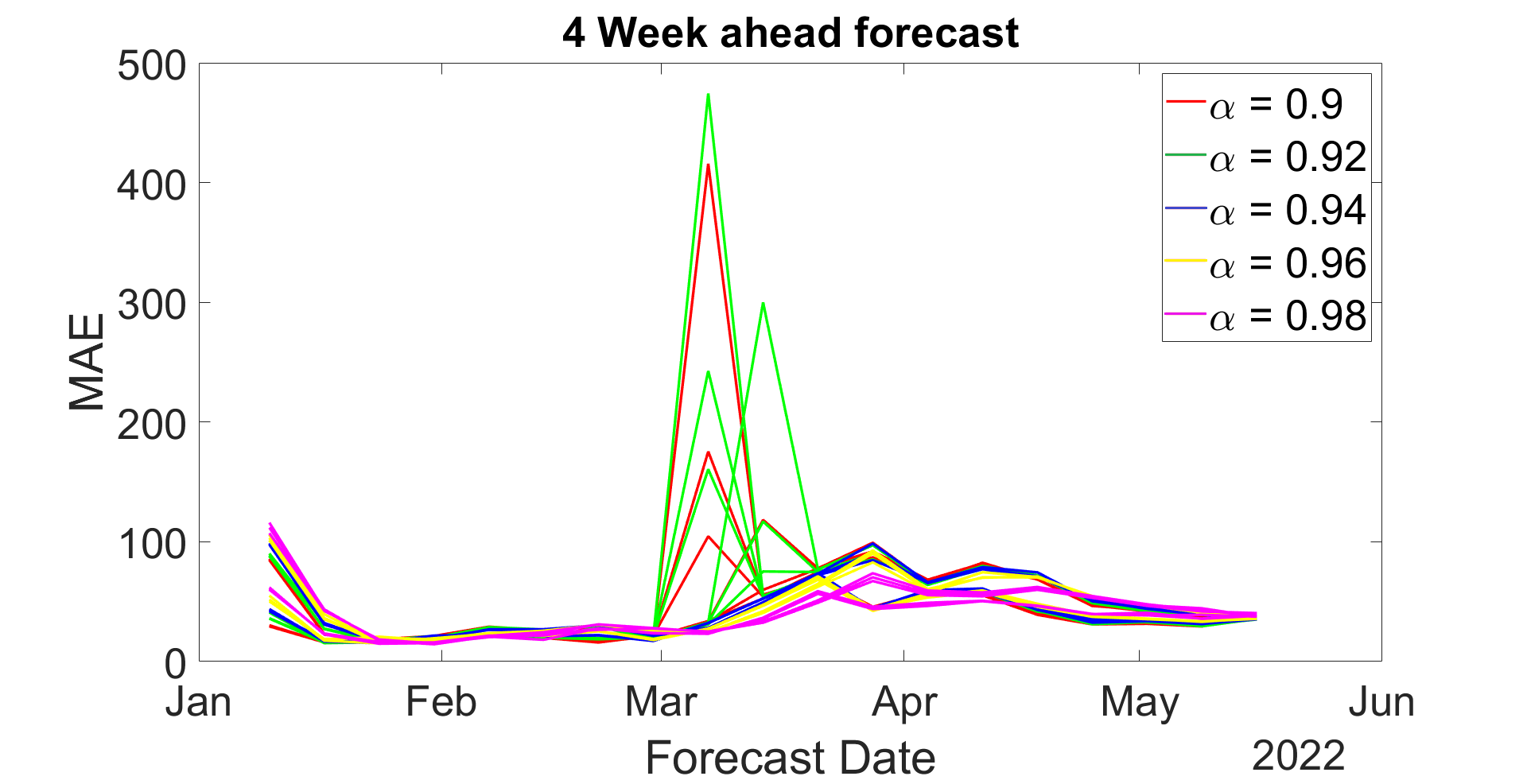}
     \caption{Sensitivity to the exponential weight $\alpha$ in weighted regression}
     \label{fig:alphaA}
     \end{subfigure}
     \begin{subfigure}{\columnwidth}
     \includegraphics[width=\columnwidth]{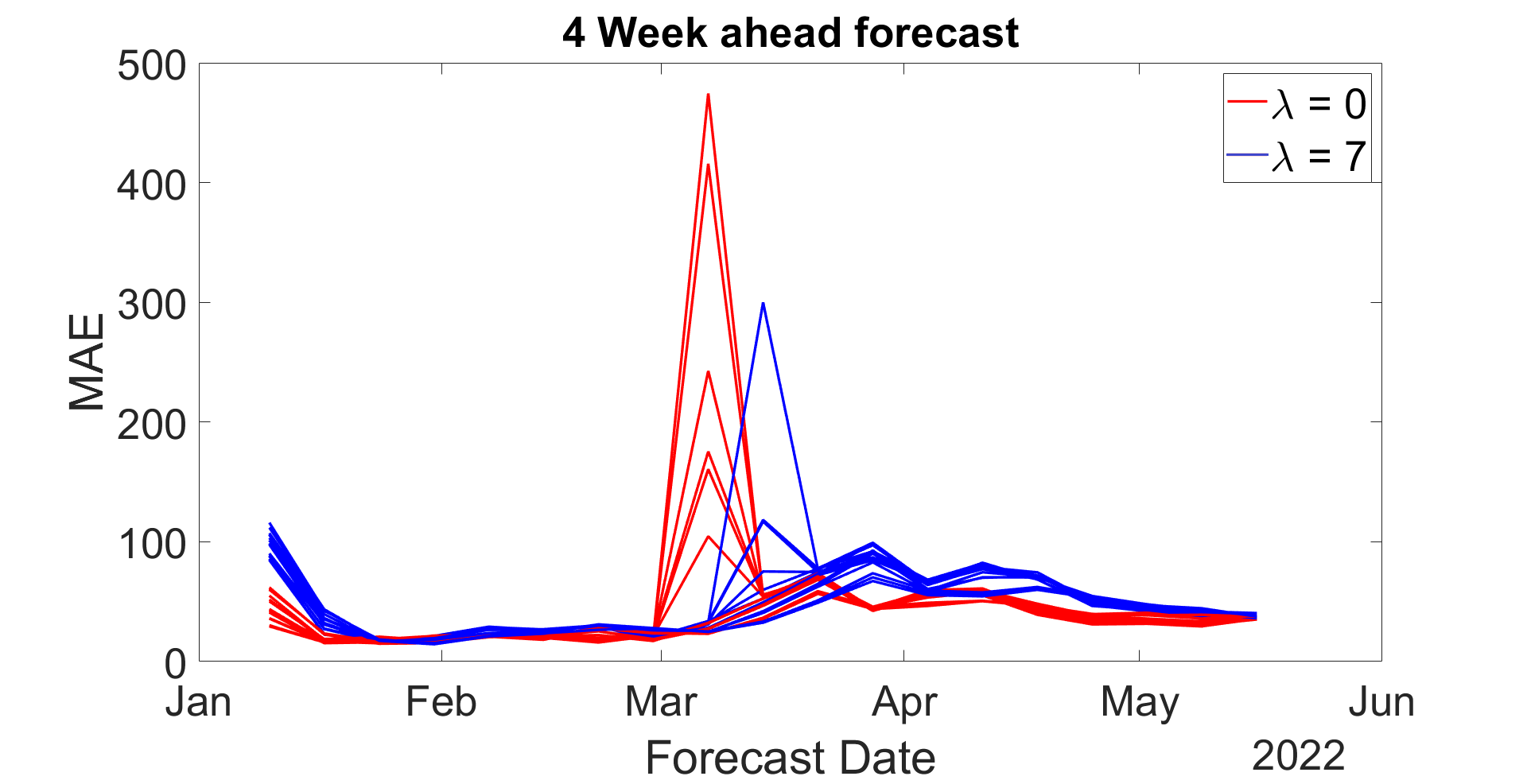}
     \caption{Sensitivity to using rates from current week ($\lambda=0$) vs last week ($\lambda=7$)}
     \label{fig:rlagA}
     \end{subfigure}
 \caption{Hyperparameter sensitivity tests for the 4 week ahead RandomForest model}
 \label{fig:hyperparam}
\end{figure}

Decision trees prioritize their splits in a manner which minimizes
the entropy (gives the largest information gain). We can use this property when analyzing the decision trees to study which features are prioritized for the splits. This technique acts as a hyperparameter sensitivity analysis since a feature with a prioritized split gives a large information gain meaning that it has a larger impact on the final result.  Figure~\ref{fig:hyperparam} shows the hyperparameter sensitivity tests by examining the individual predictors fed into the RandomForest. Each color in the plots corresponds to a different constant value of the hyperparameter being studied. We observe that $\gamma$ and $\alpha$ have a larger impact on the final result, with lower values of alpha giving a larger error some weeks. Figure~\ref{fig:DTSplits} shows the top portion of a representative sample decision tree from the RandomForest model. The prioritized splits tend to be biased towards the predictors with different values of $\gamma$ and $\alpha$, meaning that these features (or hyperparameters) give a larger information gain and are thus more sensitive.

\begin{figure}[!ht]
\centering
 \includegraphics[width=0.5\textwidth]{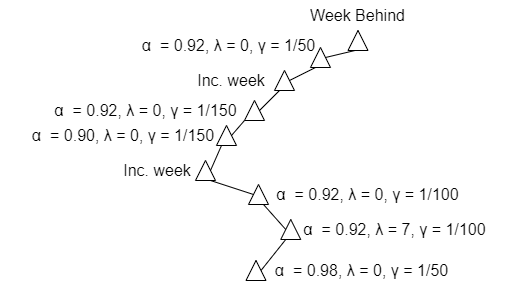}
 \caption{1 week ahead splits}
 \label{fig:DTSplitPath1}
\end{figure}

\begin{figure}[!ht]
\centering
 \includegraphics[width=0.5\textwidth]{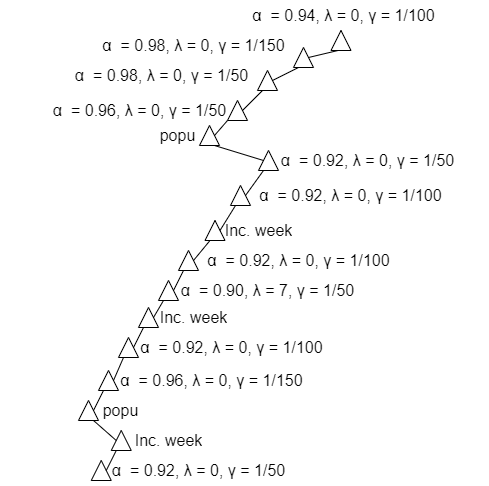}
 \caption{4 week ahead splits}
 \label{fig:DTSplitPath4}
\end{figure}

We also analyze and compare sample trees from the different week ahead RandomForest models. An observation here is that longer horizon forecasts (like 4 weeks ahead) tend to have more splits than a shorter horizon (1 week ahead) before reaching a final node. This is illustrated in figures~\ref{fig:DTSplitPath1} and figure~\ref{fig:DTSplitPath4} by observing the node statistics and following the paths along the decision trees to the final node which classifies most of the data points. We can observe that each week ahead model adapts to the task at hand and only uses as many features as required, with the 4 week ahead model requiring more features to make a better decision. This likely attributes to the RandomForest outperforming SIkJalpha and its individual predictors.

\subsection{Past Seasons: Retrospective IL Forecasts}\label{sec:past_res}
\begin{table}[!ht]
\centering
 \includegraphics[width=\columnwidth]{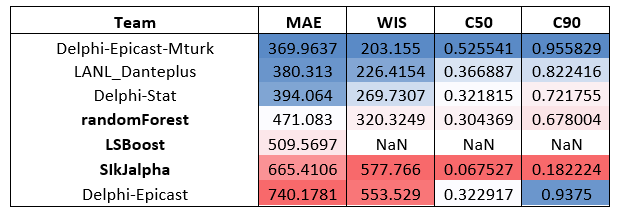}
\caption{2019 Mean evaluations over all targets with heatmap. Blue is better, red is worse.}
\label{tab:2019mean}
\end{table}

\begin{table}[!ht]
\centering
 \includegraphics[width=\columnwidth]{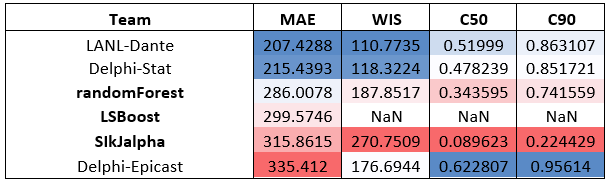}
\caption{2018 Mean evaluations over all targets with heatmap. Blue is better, red is worse.}
\label{tab:2018mean}
\end{table}

\begin{table}[!ht]
\centering
 \includegraphics[width=\columnwidth]{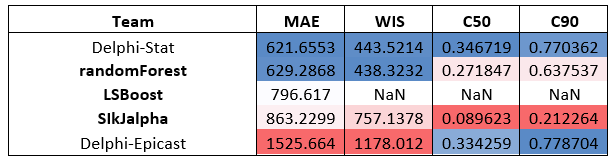}
\caption{2017 Mean evaluations over all targets with heatmap. Blue is better, red is worse.}
\label{tab:2017mean}
\end{table}

We evaluated the proposed approach on the previous seasons' challenges for ILI forecasting using the ILINET dataset (2010-2019). Tables~\ref{tab:2019mean},~\ref{tab:2018mean}, and~\ref{tab:2017mean} show the mean performance of our approaches compared to the team submissions whose approaches ranked highly based on MAE (other models have been omitted). We observe that the Random Forest is able to improve upon SIkJalpha model as before and also performed better than the LSBoosted ensemble across all seasons.

We observe that although our Random Forest does not outperform the best performing methods in the previous three seasons, its performance is on par with the other submissions, ranking in the top 50\% in 2019, 60\% in 2018, and 15\% in 2017. SIkJalpha (the mean of all variations) has a relatively low performance in the previous seasons, thus providing under-performing predictors which are likely attributing to the unsatisfactory performance of the RandomForest ensemble. This could be like due to the fact that the SIkJalpha model was modified for hospitalization forecasting based on Influezna datasets rather than ILI datasets. There are also some seasonal properties, that we assumed to be the same as the current season (2022), which likely affected the results, such as the set of choices for our uncertainty values.

\subsection{Ablation Studies}
A series of ablation studies were carried out on this season and the past seasons to study the affect of our input choices. RandomForest-SIKJalpha is the approach we have taken and discussed previously. "Predictors" feeds only the predictors to the RandomForest without utilizing the incident week, previous week, and population data points. "Data points" only provides the passed 2 weeks of hospitalization and population data to the RandomForest.
We observe in Table~\ref{tab:my_label22} that while the predictors and data points on their own are sufficient, the RandomForest-SIKJalpha is able to improve upon them in the current season's Influenza forecasting.

As for the past seasons, we notice that using the data points on their own are sufficient and significantly outperform RandomForest-SIKJalpha, increasing its ranking to the top 30\% in 2019, 35\% in 2018, and the top performing model in 2017. We also notice that RandomForest-SIKJalpha outperform the Predictors. This reaffirms our earlier statement about the predictors under-performing for ILI forecasting.

These studies show that the RandomForest is a simple yet powerful model for forecasting and can further utilize the predictors of a well tailored mechanistic model to produce good performing forecasts.

\begin{table}[]			
    \centering
    \begin{tabular}{c|c|c|c|c}
          & MAE & WIS & cov\_50 & cov\_90\\ \hline
         RandomForest-SIKJalpha & \textbf{14.1328} & \textbf{9.6354} &	\textbf{0.6504} & \textbf{0.9665}\\
         Predictors & 17.3929 & 11.8047 &0.6169 &	0.9507\\
         Data points& 18.0286 & 11.8226 &	0.6499 &	0.9633
    \end{tabular}
    \caption{Ablation study 2022}
    \label{tab:my_label22}
\end{table}

\begin{table}[]			
    \centering
    \begin{tabular}{c|c|c|c|c}
          & MAE & WIS & cov\_50 & cov\_90\\ \hline
         RandomForest-SIKJalpha & 471.0830 & 320.3249 &	0.3044 & 0.6780\\
         Predictors & 474.5512 & 321.0961 &0.3766 &	0.7584\\
         Data points& \textbf{423.1405} & \textbf{290.1701} &	\textbf{0.3692} & \textbf{0.7587}
    \end{tabular}
    \caption{Ablation study 2019}
    \label{tab:my_label19}
\end{table}
	
\begin{table}[]			
    \centering
    \begin{tabular}{c|c|c|c|c}
          & MAE & WIS & cov\_50 & cov\_90\\ \hline
         RandomForest-SIKJalpha & 286.0078 & 187.8517&	0.3436 & 0.7416\\
         Predictors & 291.2943 & 195.4995 & 0.3500 &	0.6691\\
         Data points& \textbf{246.3770} & \textbf{168.8018} &	\textbf{0.4148} & \textbf{0.8046}
    \end{tabular}
    \caption{Ablation study 2018}
    \label{tab:my_label18}
\end{table}

\begin{table}[]			
    \centering
    \begin{tabular}{c|c|c|c|c}
          & MAE & WIS & cov\_50 & cov\_90\\ \hline
         RandomForest-SIKJalpha & 629.2868&	438.3232&	0.2718&	0.6375\\\
         Predictors & 681.4325&	471.9884&	0.3443&	0.7535\\
         Data points& \textbf{565.0680}&	\textbf{394.4054}&	\textbf{0.3515}&	\textbf{0.7562}\\
    \end{tabular}
    \caption{Ablation study 2017}
    \label{tab:my_label17}
\end{table}

\subsection{Implementation}
The methods were implemented in MATLAB on an 6-core [intel(R) Core(TM) i7-8750H CPU@ 2.20GHz] machine with 20 GB RAM. The code including integrating data resources, generating predictors, training, inferences, and forecast submission are publicly available~\footnote{\url{https://github.com/maa989/RandomForest-SIkJalpha}}. An advantage of our approach is its speed making it easily scalable to multiple regions. The runtime of different components are given in Table~\ref{tab:Runtime}. The most time-consuming aspect is to generate training data for Random Forest. This involves generating predictions from $30$ variations of SIkJalpha for all states from all past seasons. However, this computation is only performed once. We save the results, and during the current season, every week we generate new predictions from SIkJalpha models with the new data and append with the saved data which takes only around 14s. Further, while we report runtime of serial implementation, generating predictions for the past season is embarrassingly parallel as each week, each state, and each variation of SIkJalpha can be executed independently.
\begin{table}
    \centering
    \begin{tabular}{c|c|c}
    \hline
          & Runetime (s) & Frequency\\
          \hline
         SIKJalpha pred history & 4499 & once\\
         SIKJalpha pred current  & 14.4 & weekly\\ 
         Random Forest training &  152.6 & weekly\\
         Random Forest inference & 80.1 & weekly
    \end{tabular}
    \caption{Approximate runtimes of various components of the script.}
    \label{tab:Runtime}
\end{table}

\subsection{Future Work}

We will use this approach in the future Influenza seasons and for other infectious diseases around the world. Our team also participates in several forecasting and scenario projection efforts for COVID-19~\cite{lab_covid-19_2020,cramer_evaluation_2021,europe_forecast_hub,SMH,ESMH,bracher_short-term_2020}. Currently, we use variation of SIkJalpha model with various complexities, including multi-variants dynamics and multiple susceptibility states (combinations of prior infections and doses of vaccines). With reduction in case reporting, increased home tests, future doses and combinations of more transmissible and immune escape variants, capturing the exact dynamics from the data is becoming more difficult. We envision that our Tree Ensmeble-based approach with simple mechanistic models as described in this paper, may be useful in the future of COVID-19 hospitalization and death forecasting.  We will also explore other interpretable machine learning methods. Random Forest allowed us to easily generate the desired quantiles. We will explore how to generate quantiles for other approaches.


\section{Conclusions}
We proposed combining simple mechanistic models and Tree Ensembles for improved forecasting of Influenza hospitalizations. Our approach does not rely on any human intervention or manual tuning for particular states and time. We extended the SIkJalpha model to predict Influenza hospitalizations and generated variations of this model based on various combinations of hyper-parameter choices. 
We demonstrated that the Tree Ensemble is able to produce lower errors when compared to the mean of the variations of SIkJalpha and the individual predictors. In the past Influenza seasons, retrospective evaluation demonstrated that our approach is on par with other submissions. We participated in FluSight 2022 challenge with our approach to produce prospective forecasting. Our approach was able to produce good performance outperforming all other methods, including FluSight-Ensemble, despite it being fully automated, without any manual tuning per state or per week.

\section*{Acknowledgements}
This material is based upon work supported by the National Science Foundation under Grant No. 2135784 and the University of Southern California Strategic Directions Research Award.

  \bibliographystyle{ACM-Reference-Format}
  \bibliography{sample,epidemic,my_pubs}

\end{document}